# HAGCN : Network Decentralization Attention Based Heterogeneity-Aware Spatiotemporal Graph Convolution Network for Traffic Signal Forecasting

JunKyu Jang, SungHyuk Park

**Abstract**— The construction of spatiotemporal networks using graph convolution networks (GCNs) has become one of the most popular methods for predicting traffic signals. However, when using a GCN for traffic speed prediction, the conventional approach generally assumes the relationship between the sensors as a homogeneous graph and learns an adjacency matrix using the data accumulated by the sensors. However, the spatial correlation between sensors is not specified as one but defined differently from various viewpoints. To this end, we aim to study the heterogeneous characteristics inherent in traffic signal data to learn the hidden relationships between sensors in various ways. Specifically, we designed a method to construct a heterogeneous graph for each module by dividing the spatial relationship between sensors into static and dynamic modules. We propose a network decentralization attention based heterogeneity-aware graph convolution network (HAGCN) method that aggregates the hidden states of adjacent nodes by considering the importance of each channel in a heterogeneous graph. Experimental results on real traffic datasets verified the effectiveness of the proposed method, achieving a 6.35% improvement over the existing model and realizing state-of-the-art prediction performance.

**Index Terms**— graph neural network, graph convolution network, traffic forecasting, spatiotemporal network

—————————— ◆ ——————————

## 1 INTRODUCTION

Time series forecasting is crucial for dynamically predicting real-world phenomena. Among them, traffic forecasting is an important application of this technique for real-world scenarios. Traffic forecasting aims to predict future traffic conditions based on historical traffic data. Because accurate traffic forecasting can be highly beneficial for controlling urban traffic [1, 2], researchers have used deep learning methods to develop traffic forecasting. In particular, because traffic conditions in neighboring areas may affect each other, a graph convolution network (GCN) is a widely used method to adequately capture spatial correlations for traffic forecasting [3]. Recent studies have focused on spatiotemporal graph modeling to maximize GCN utilization. For example, DCRNN [4] models spatial dependencies between nodes using a bidirectional graph random walk, and captures temporal relationships with a recurrent neural network. In addition, GraphWaveNet [5] and GMAN [6] have attempted to elaborate spatial-correlation training using adaptive graphs. Furthermore, models such as the DMSTGCN [7], ACRGN [8], and Z-GCNETs [9] have been proposed to model the dynamic spatial

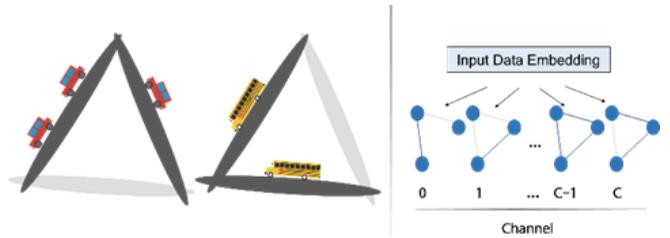

Fig. 1. Heterogeneous spatial correlation. (a) A road network that has a variety of unknown relationships as features measured by sensors. (b) Transform the input data into c channels and capture the unknown relation using a heterogeneous graph.

dependency between traffic nodes.

Despite efforts to accurately learn spatial dependencies between traffic sensors, there has been little discussion of how to learn relationships between hidden channels that can be extracted from sensors. Traffic sensors typically only capture one or two functions, such as traffic speed or traffic volume. In most existing studies, one or two traffic signal data go through the process of embedding as a hidden channel to learn hidden patterns. But the spatial dependency for each channel of the embedded traffic data is considered to be the same and has not been trained.

However, hidden patterns of traffic signals can imply different types of information, and the spatial correlation between traffic sensors for each hidden pattern is completely different. For example, suppose that the sensor on the upper left road in Fig. 1(a) captured the average traffic speed of the road. However, there are different types of vehicles on the road, such as public transport and private cars, which are hidden patterns for road speed, and they work

────────────────────────────
• *JunKyu Jang and SungHyuk Park are with the KAIST College of Business , 85, Hoegi-ro, Dongdaemun-gu, Seoul, Republic of Korea. E-mail :{jbkjsm, sunghyuk.park}@kaist.ac.kr*

together to create traffic speeds. In Fig. 1(a), the traffic patterns generated by public transport and private cars on upper left road are spatially related to the other roads below and to the right, respectively. This is because the roads mainly used by the two types of vehicles are different. As in previous papers, if only one spatial correlation between sensors is learned from the data accumulated by the sensor, different spatial correlations hidden in the data cannot be learned like other spatial relationships hidden in public transportation and private cars in Fig. 1(a). Therefore, it is necessary to discover and learn these hidden relationships in various ways. We hypothesize that learning the spatial relationships of potential patterns of traffic signals separately for each hidden channel as shown in Fig. 1(b) can help predict traffic signals.

In this process, a few problems may exist because of the complexity of accurately designing the spatial relationship of the hidden channels extracted from the nodes. The spatial relationships between nodes are not uniformly dynamic or static. To accurately reflect the spatial relationship between nodes, one must learn the dynamic and static relationships separately. In addition, even if a heterogeneous graph model of the relationship between the hidden features is constructed, the importance of each hidden channel cannot be reflected by the existing GCN method.

In this study, we propose a heterogeneity-aware GCN for traffic signal prediction using a heterogeneous graph. First, the heterogeneous graph generator aims to accurately model the spatial dependency between sensors using Tucker decomposition. The spatial relationship between each sensor is modeled as a three-dimensional (3D) and four-dimensional (4D) tensor by dividing it into two adjacency matrices that capture static and dynamic relationships. In addition, we propose a heterogeneity-aware channel-attention GCN to reflect the importance of hidden channels in learning spatial relationships through graph convolution. And when we perform channel attention, we conceived of a network decentralization attention module based on network analysis. Attention using network decentralization pooling is modeled to give attention to a hidden channel network that utilizes information from various nodes. In our model, several parallel structures are used to capture various spatiotemporal relationships, and each layer is composed of a block of the proposed heterogeneous graph structure and a heterogeneous channel-attention GCN. The traffic signal was transformed using temporal convolution networks (TCNs) to capture various temporal attributes, and the final traffic signal prediction was obtained using skip connections at each layer.

The main contributions of this study are summarized as follows:

•**Method.** We proposed HAGCN, a learning method for spatial correlation using a heterogeneity-aware graph. HAGCN learns two relationships by designing a static/dynamic generator to accurately learn spatial correlation. We show that this static/dynamic division of spatial correlation is effective in predicting traffic signals.

•**Theory.** We provide a heterogeneity-aware GCN that applies channel attention inspired by network decentralization to learn the spatial correlation of hidden channels. We formulate the concept of network decentralization of traditional network analysis and propose a method to give attention to hidden channels that utilize information from various nodes.

•**Experiments.** we perform extensive experiments on real traffic data. HAGCN shows state-of-the-art performance in predicting various time intervals for various types of traffic data such as traffic speed and traffic volume prediction. We also verified our proposed channel attention method and static/dynamic module. Finally, a robustness check was performed on noisy data.

## 2 RELATED WORKS

### 2.1 Time Series and Traffic Prediction

Time series forecasting has been studied for several decades, and traffic prediction is one of the most active research areas in time series forecasting. Traditionally, time-series forecasting has been developed using statistical-based methods such as ARIMA [10] and VAR [11]. However, these methods have limitations because they are based on the assumption that the time series is stationary and has a linear combination with the time lag variable. Many researchers have recently used deep learning approaches to represent complex time series relationships. As deep learning methods have emerged as the dominant methodology, recurrent neural networks (RNNs) have become popular for time series prediction. In particular, long short-term memory (LSTM) [12] and gate recurrent unit (GRU) [13] architectures for time-series prediction were highlighted by partially solving the instability of the RNN by exploiting the concept of flexible learning based on the memory and forget gates. Subsequently, TCNs using convolution networks were proposed separately from the RNN series method to predict time series. TCN-based algorithms have also been widely used for time series predictions [14, 15].

In the case of traffic forecasting, researchers have attempted to consider actual spatial (sensor location) correlations between traffic time series, in addition to the temporal correlations that are generally considered. In these studies, a GCN-based model was used to capture the spatial correlation of regions close to each other based on the assumption that a traffic time series is generated for a specific location [4, 16]. Although GCN-based research has achieved state-of-the-art traffic forecasting in recent years, most studies have constructed spatial TCNs using a predefined spatial graph structure.

### 2.2 Spatio-Temporal Graph Neural Networks

Advancements in graph neural networks and time series data learning methods have made it possible to perform spatiotemporal learning. In contrast to the relatively simple application of GCNs to train networks using spatial graph structures, researchers have attempted to construct networks that appropriately execute spatio-temporal

learning [5, 8, 9, 17, 18].

In order to capture complex spatial temporal relation between traffic data using GCN, it was essential to accurately check the relationship between nodes by properly learning the adjacency matrix rather than predefining it as in the existing method. Among them, GraphWavenet [5] used a GCN in the spatial domain and 1D convolution along the time domain to separately learn the adjacency matrix and time series. STSGCN [18] proposed a method for capturing local correlations through a localized synchronous spatiotemporal graph convolution module independent of global mutual effects. LSGCN [19] generates features using a novel method called cosine graph attention in spatially gated blocks. AGCRN [8] learns the spatiotemporal correlation for each node by separating the parameters into spatial and temporal components using node-adaptive parameter learning. SFTGNN [20] effectively captures hidden spatial dependencies by fusing a spatial graph with a temporal graph. STNN [21] also designed a novel space time module to learn the local spatio-temporal correlations to attempts to capture spatial and temporal patterns well.

Z-GCNETs [9] incorporate a zigzag topological layer into the GCN. DMSTGCN [7] is used to generate a dynamic graph and construct a framework that uses auxiliary features separately. In STG-NCDE [22], two neural-controlled differential equations for spatiotemporal processing were constructed. Although these methods have significantly contributed to the development of spatiotemporal learning, learning about hidden channels of traffic signals that are essential in spatiotemporal network construction has not been considered. Unlike existing methods, this study proposes a novel heterogeneity-aware and dynamic learning module to accurately capture spatio-temporal relationships for each hidden channel of a traffic signal. Our model is designed to precisely capture the attention of the network for each hidden channel by using the channel attention technique appling network decentralization pooling. In addition, two different modules that learn static and dynamic patterns were created. By fusing and integrating the two types of patterns, the spatial relation between traffic signals is learned.

## 3 PRELIMINARIES

This section describes the heterogeneous graph and the notation of the traffic features used in this study. Subsequently, we formulated the problem of predicting traffic signals using historical traffic features.

### 3.1 Time Series and Traffic Prediction

In this study, we assume that the traffic network has a dynamic characteristic that varies according to time $t$ and heterogeneous hidden embedding. We define a heterogeneous dynamic traffic network as follows: A heterogeneous dynamic graph is defined as a graph $G_t = (V, E_t, J_t)$ associated with an edge-type mapping function $\psi: E_t \to J_t$, where V denotes the set of nodes, $E_t$ is the set of edges, and $J_t$ denotes sets containing the weights of connected edges.

In practice, a node may represent a detection sensor located on a road in a traffic network. Each node records traffic flow data such as vehicle speed and traffic volume at regular intervals. Therefore, at time step $t$, graph $G_t$ has a feature matrix $X_t \in R^{N \times d}$ where $d$ is the input feature dimension, and N is the number of nodes. Given the graph signals $X_t$, we pass a linear layer through $X_t$ to create $F$ hidden channels and embed the traffic signal. Therefore, we aim to learn adjacency matrix $A \in R^{F \times T \times N \times N}$ with F heterogeneity properties. $A$ is a 4D tensor, and the elements $A_{f,t,i,j}$ denote the edge weights of feature $f$ at time $t$ of vertices $v_i$ and $v_j$. Furthermore, at time $t$, the 3D adjacency matrix in graph $G_t$ is defined as $A_t$.

### 3.2 Problem Definition

Traffic forecasting aims to predict future traffic signals based on historical traffic features. At time step $t$, given the adjacency matrix $A_t$ and historical graph signals of $P$ steps, we want to learn mapping function $f$ to predict the graph signal of the next step $Q$. This can be defined as

$$[X_{t-P+1:t}, A_t] \xrightarrow{f} X_{t+1:t+Q} \quad (1)$$

where $X_{t-P+1:t} \in R^{P \times N \times d}$ and $X_{t+1:t+Q} \in R^{Q \times N \times d}$.

## 4 METHODOLOGY

In this section, we introduce our proposed model in detail. First, we explain the construction method of a heterogeneous graph that separately captures the correlation between sensors for static and dynamic graphs. Next, we describe the network decentralization channel attention based heterogeneity-aware GCN used to learn spatiotemporal dependencies while capturing the relative importance of the hidden channels. Finally, we briefly describe the architecture of the framework.

### 4.1 Heterogeneity-Aware Graph Constructor

The spatial correlation between traffic sensors changes over time owing to the dynamic nature of traffic. Because the spatial correlation between hidden channels of traffic signals may vary across features, it is essential to properly design the adjacency matrix of the graph. In previous studies, the spatial correlations between sensors were predefined using a two-dimensional matrix. Although some studies have considered dynamic correlation for learning, the homogeneity-aware graph does not consider the heterogeneity of correlations between unknown hidden traffic channels. In this study, the relationship between the nodes of a heterogeneous traffic signal is captured by dividing it into dynamic and static relationships, regardless of time. We divide the 4D adjacency matrix $A$ into two matrices: the 4D matrix $A^d$, which captures the dynamic spatial relationship, and the 3D matrix $A^s$, which captures the static spatial relationship, and define it as follows.

$$A_{f,t,i,j} = A^d_{f,t,i,j} + A^s_{f,i,j} \quad (2)$$

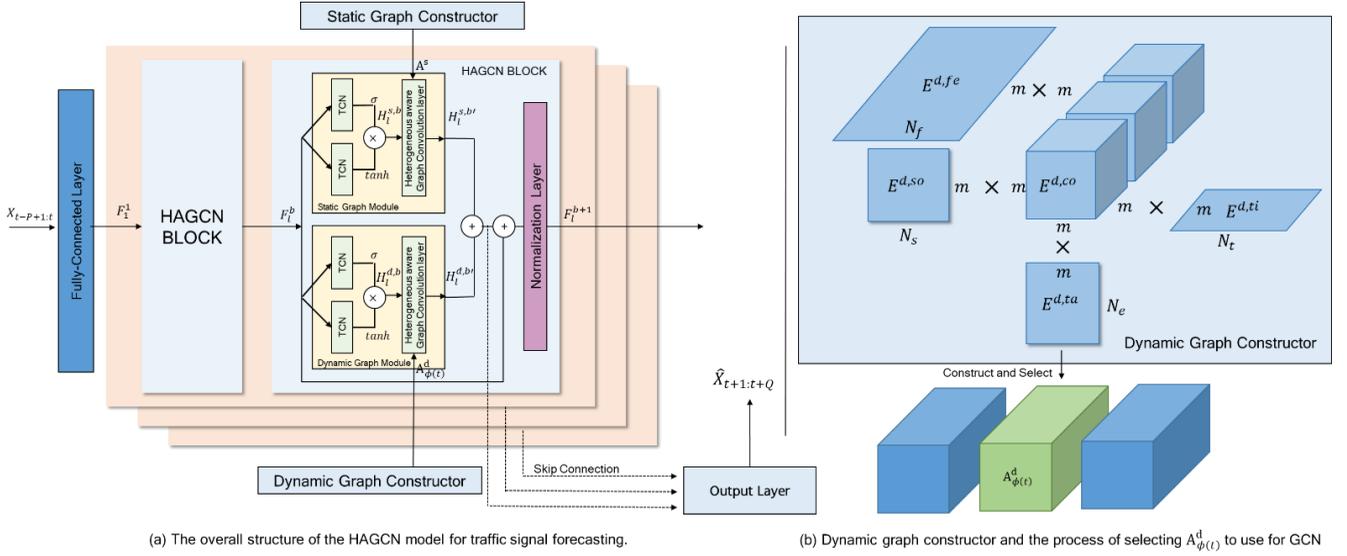

Fig 2. Network architecture of HAGCN. The HAGCN block of the framework consists of parallel parts that learn two spatial correlations of static/dynamic modules. Each static/dynamic module consists of a part that learns temporal correlation as a temporal convolution layer and a heterogeneity-aware graph convolution layer that learns a graph generated by a generator. For the final prediction, the ouput captured by the last block's graph convolution of each layer is taken and used for prediction

where $A^d \in R^{F \times T \times N \times N}$ captures the dynamic relationship between sensors as a 4D tensor, and $A^s \in R^{F \times N \times N}$ captures the static relationship between the sensors as a 3D tensor. where F is the number of hidden channels, N the number of nodes, and T the number of time intervals. $A^d$ and $A^s$ were utilized through graph convolution in the dynamic and static modules of the heterogeneity-aware channel attention graph convolution network (HAGCN) model network, respectively.

The easiest way to learn these two adjacency matrices is to determine the relationship between nodes by directly assigning parameters to all elements of the dynamic and static matrices. However, with this assignment, the computational complexity of the adjacency tensor is $O(FTNN + FNN)$, If different dynamic correlations are learned for all time signals of the train traffic signal, not only is the dynamic correlation undetermined when applied to the validation and test sets, but the number of parameters that need to be learned increases in proportion to the number of training time intervals. Therefore, we assume that traffic signals have a certain periodicity and that traffic signals detected during the same time of day share the same dynamic correlation.

$$A^s_{f,i,j} = max(0, \sum_{a=1}^{m}\sum_{b=1}^{m}\sum_{c=1}^{m} E^{s,co}_{a,b,c} E^{s,ch}_{f,a} E^{s,ta}_{i,b} E^{s,so}_{j,c}) \quad (3)$$

$$A^d_{f,t,i,j} = \sum_{a=1}^{m}\sum_{b=1}^{m}\sum_{c=1}^{m}\sum_{d=1}^{m} E^{d,co}_{a,b,c,d} E^{d,ch}_{f,a} E^{d,ti}_{t,b} E^{d,ta}_{i,c} E^{d,so}_{j,d} \quad (4)$$

$$A^d_{f,t,i,j} = max(0, A^{d'}_{f,t,i,j}) \quad (5)$$

In the above equation, $E^{s,ch}$, $E^{d,ch} \in R^{N_f \times m}$ are the embeddings of the hidden channel; $E^{s,ta}$, $E^{d,ta} \in R^{N_e \times m}$ are the embeddings of the target node; $E^{s,so}$, $E^{d,so} \in R^{N_s \times m}$ are the embeddings of the source node; and $E^{s,co} \in R^{m \times m \times m}$ and $E^{d,co} \in R^{m \times m \times m \times m}$ represent the core tensor; $E^{d,ti} \in R^{N_t \times m}$ is the embedding of the time interval; $N_f$ is the number of hidden features; $N_t$ is the number of time intervals within one period; $N_s$ and $N_e$ are the numbers of original and target nodes, respectively; and m is the embedding dimension to reduce the dimension.

we can train 3D tensor $A^s$ and 4D tensor $A^d$ with large number of parameters by training only E which is the embedding vectors of tucker decomposition. In fact, traffic signal relationships hide many repeating patterns. Information concerning duplicate patterns can be filtered via dimensionality reduction using the Tucker decomposition.

## 4.2 Heterogeneity-Aware Channel Attention GCN

The spatial correlation of the hidden channels of different nodes can contribute to improving the prediction performance of the traffic signal that we want to predict. In previous studies, graph convolution has been used to extract and utilize features unique to each node by aggregating the spatial correlation and signals of neighboring nodes. However, in this study, we propose a dynamic-heterogeneity-aware GCN to model the heterogeneous properties of the hidden features of nodes with different spatial dependencies for each channel and the dynamic correlation of nodes.

First, the heterogeneity-aware graph convolution method, which aggregates and updates the hidden state of nodes in each learning step using the 3D adjacency matrix $A'$ representing the heterogeneous graph, is defined as:

$$H_l^{b'} = \sum_{k=0}^{K} (\phi_k(A') \circledcirc (A')^k) H_l^b \theta_k \quad (6)$$

where $K$ is the maximum diffusion step, $H_l^b$ is the temporal convolution output of the $b$-th block of the $l$-th layer, $\circledcirc$ is the element-wise product, and $\theta_k$ is the convolution parameter of the k-th diffusion step. $A'$ is the adjacency matrix used for graph convolution.

In our model, the adjacency matrix uses $A^d$ to capture dynamic relationships and $A^s$ to capture static spatial dependencies. First, when generating and using the adjacency matrix $A^s$ in the GCN in the static module, $A^s$ is substituted for $A'$ and used for graph convolution. However, when using $A^d$ in the graph convolution in the dynamic module to learn dynamic relationships, graph convolution is used by considering only $A_{h(t)}^d$, which corresponds to the dependency of time $t$ in $A^d$. Using the function $h(t)$ to calculate the number of time intervals from the period at time $t$, we utilize 3D tensor $A_{h(t)}^d$.

In this study, the channel attention function $\phi_k(A')$ is used in the adjacency matrix $A'$ to assign weight to the hidden channel using spatial correlation with more neighbor nodes. In particular, the importance of channels in the 3D heterogeneity-aware adjacency matrix differs. For example, because one hidden channel has a frequent and large spatial correlation with neighboring nodes, many edge weights are greater than zero. Such a channel with frequent correlation can be considered valuable in utilizing graph convolution, which is advantageous for the use of information from the neighboring node when an edge weight appears. In addition, because we argue that channels in the adjacency matrix influence each other, we designed an attention module using a layer that calculates the interaction between channels. Consequently, channel attention $\phi_k(A')$ is defined as follows :

$$\phi_k(A') = \sigma(f_{\{w_1, w_2\}}(g(A'))) \quad (7)$$

where σ is a softmax function and $g(A')$ is a channel-wise aggregate function. In general, in the channel attention module in the field of computer vision, each channel means one image convolution map. Therefore, channel attention is calculated for each channel using methods such as global average pooling or generalized mean pooling as a channel-wise aggregate function to give attention according to the activation level of the entire channel pixel [10, 11]. However, since a channel in a heterogeneous graph convolution network means a network of each hidden channel, the aggregate function must be tuned differently. Network decentralization is a measure of how non-centralized the most central node is compared to all other nodes [12]. Therefore, if the network decentralization is high, it means that the graph is organically connected with various nodes. We define the network decentralization function $g(A')$ as follows to give attention to hidden channels that actively utilize

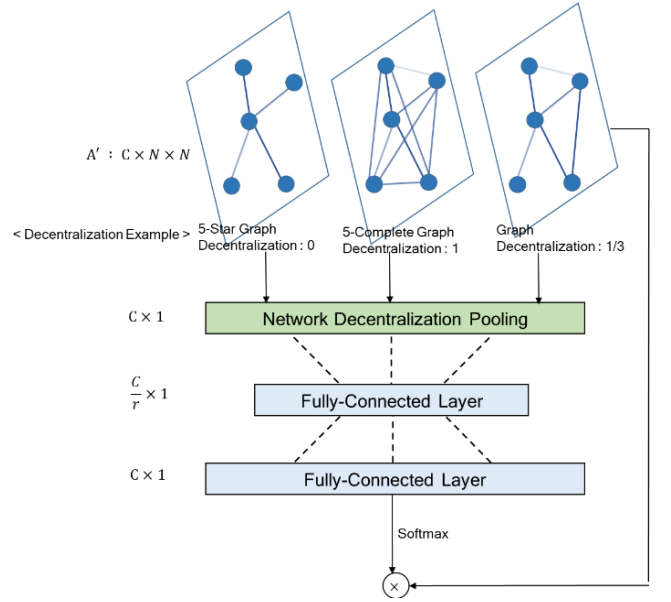

Fig 3. Network decentralization based channel attention method

information from various nodes when performing graph convolution.

$$g(A')[c] = 1 - \frac{\sum_{v \in A_c'} \left( M_{A_c'}(\dot{v}) - M_{A_c'}(v) \right)}{\max_{\forall G \text{ has } N \text{ node}} \sum_{v \in G} (M_G(\dot{v}) - M_G(v))} \quad (8)$$

where $c$ is the each channel, and $g(A')[c]$ is the $c$-th element of the aggregate function. $M_G$ is node centrality function of graph G. We use degree centrality that means the sum of degrees of nodes among various node centrality functions to calculate network decentralization. $\dot{v}$ stands for argmax $M_G$ which is the vertex with the highest degree centrality in graph $G$. In order to have the maximum value of the decentralization (8)'s denominator, it should be an N-star graph in which all edges are gathered at one node. Therefore, when calculating (8) considering degree centrality as follows:

$$g(A')[c] = 1 - \frac{N \times \max_{\forall i}(\sum_{j=1}^{N} A'_{c,i,j}) - \sum_{i=1}^{N} \sum_{j=1}^{N} A'_{c,i,j}}{(N-1)(N-2) \times \max_{\forall i,j}(A'_{c,i,j})} \quad (9)$$

Let $y = g(A'), y \in F \times 1$, and F is the number of hidden channels. $f_{\{w_1, w_2\}}$ takes the form

$$f_{\{w_1, w_2\}}(y) = W_2 \max(0, W_1 y) \quad (10)$$

To avoid the high complexity of the attention module, the sizes of $W_1$ and $W_2$ were designated as $F \times \frac{F}{r}$ and $\frac{F}{r} \times F$, respectively [10]. $f_{\{w_1, w_2\}}$ uses one fully connected (FC) layer to project the output feature of the pooling layer to a low-dimensional space. Subsequently, it is re-mapped to indirectly create a correspondence between the output feature of the pooling layer and attention weight of the adjacency matrix channel.

## 4.3 Temporal Convolution Layer

The traffic signal of a node is highly correlated with its historical data. Dilated causal convolution with a TCN was used to capture the temporal trends of the nodes. Dilated causal convolution increases the depth of the TCN layer, thereby enabling exponentially larger receptive fields. Unlike RNN-based time-series approaches, dilated causal convolution networks adequately process sequences over long distances in a non-recursive manner. Therefore, dilated causal convolution networks are suitable for data with periodicity, are easy to compute in parallel, and alleviate the problem of gradient explosion. However, the gating mechanism is important in RNN-based approaches. Their ability to control the information flow through layers in temporal convolutional networks has been demonstrated [13]. Therefore, we used the gated TCN in our model to learn complex time dependencies.

$$H_l^b = \tanh(Z_1 * F_l^b) \odot \sigma(Z_2 * F_l^b) \quad (11)$$

where $Z_1$ and $Z_2$ are model parameters, $\odot$ is the element-wise product, and $\sigma(\cdot)$ is a sigmoid function that determines the proportion of information transferred to the next layer.

## 4.4 Other Components

In the proposed model, $F_b^l$ is the input of the b-th block of the l-th layer and the output of the (b − 1)-th block of the l-th layer. The data $F_1^1$ input to the first block of the first layer are transformed by the FC layer as follows:

$$F_1^1 = W_{in} X_{t-P+1:t} + b_{in} \quad (12)$$

In this model, the output $F_{b+1}^l$ of the block is calculated by using the residual link at the end of each block. Subsequently, the sum of the outputs from the static and dynamic modules of the last block of each layer was used as the skip connection, and the concatenated value of the skip connections of each layer was sent to the output layer and used for the final prediction. The output layer was constructed using two FC layers.

$$\begin{aligned} F_{b+1}^l &= H_l^{d,b'} + H_l^{s,b'} + F_b^l \\ Skip_l &= H_l^{d,-1'} + H_l^{s,-1'}, Skip = \| Skip_l \\ Out &= W_{out1} ReLU(Skip) + b_{out1} \\ \hat{X}_{t+1:t+Q} &= W_{out2} ReLU(Out) + b_{out2} \end{aligned} \quad (13)$$

We chose the L1 loss as our training objective and optimized the loss for multistep prediction. Thus, the loss function of our model for multistep traffic prediction can be formulated as follows :

$$L(W_\theta) = \sum_{i=t+1}^{t+Q} |X_i - \hat{X}_i| \quad (14)$$

where $W_\theta$ represents all learnable parameters, $H_l^{d,-1'}$ and $H_l^{s,-1'}$ are outputs of the static and dynamic modules of the last block of layer $l$, respectively, and ∥ is the concatenate function.

## 5 EXPERIMENTS

In this section, we compare the performance of the proposed method with that of existing state-of-the-art methods using PeMSD4 and PeMSD8. Subsequently, we verify whether the proposed module and method improved the performance of the model. Finally, we test the performance of the model on noisy data using a robustness check.

### 5.1 Datasets

We conducted experiments on two public real-world traffic datasets, PeMSD4 and PeMSD8, to evaluate task performance. PeMS is the Caltrans Performance Measure System (PeMS), which measures California's highway traffic in real time every 30 s [14].

PeMSD4: The pemsd4 dataset contains traffic flow data from the San Francisco Bay area. Data from 307 sensors were collected from January 1, 2018, to February 28, 2018.

PeMSD8: The pemsd8 dataset contains traffic flow signals from the San Bernardino area. Data from 170 sensors were selected from July 1, 2016, to August 31, 2018.

### 5.2 Preprocessing

For all datasets, we filled in the missing values using the linear interpolation method and used the training set data to perform normalization using the Z-score method for training.

To facilitate the training of our model after injecting the initial edge weights into the adjacency matrices $A^d$ and $A^s$, we applied Tucker decomposition by defining the initial adjacency matrix to apply the initial parameter to $E(\cdot)$, which is the embedding parameter constituting $A^d$ and $A^s$ for training. The initial adjacency matrix is defined as follows:

$$\begin{aligned} \forall t,f \; & A_{f,i,j}^{s,(0)}, A_{t,f,i,j}^{d,(0)} \\ &= \begin{cases} \exp\left(-\dfrac{d_{i,j}^2}{\delta}\right), if \; i,j \; are \; neighbors \\ 0, if \; i,j \; aren't \; neighbors \end{cases} \end{aligned} \quad (14)$$

### 5.3 Baselines

To verify the performance of our model, we selected models that include traditional statistical methods, various graph neural network models, and methods proposed in recent key papers as benchmarks. The benchmark methods are described as follows:

- DCRNN [4]: A model for achieving multistep prediction by employing the diffusion process and

graph convolution. A diffusion-convolutional neural network was built by combining a GCN and recurrent models using the encoder-decoder method.
- GMAN [6]: A multi-attention model based on spatial and temporal embeddings of traffic signals.
- GraphWavenet [5]: A model that applies an adaptive graph network and combines dilated causal convolution and graph convolution.
- DMSTGCN [7]: A spatiotemporal network that generates a dynamic graph and introduces a framework that uses auxiliary features separately.
- AGCRN [8]: A model that uses node adaptive parameter learning to separate parameters into spatial and temporal components.
- Z-GCNETs [9]: A spatiotemporal graph that integrates a time-aware zigzag topological layer with graph convolution.
- STG-NCDE [15]: Two neural-controlled differential equations for spatiotemporal processing were constructed and studied in a new manner.

### 5.4 Experimental Setup

Our experiments were performed in a computer environment with an Intel® Xeon® Gold 5119T CPU at 1.90 GHz and four NVIDIA RTX 3090 GPU cards. In our model, to span the length of the input sequence, we used eight dilations using the dilation factor [1, 2] for each of the two blocks in the four layers. Because of the characteristics of our model, the output channel sizes of the dilated and graph convolutions were set to 32. The batch size was set to 64 and the initial learning rate was set to 0.001. The core dimensions of the Tucker decomposition were reduced to 40, and the maximum depth of the graph convolution layers was set to 2. The model was optimized using the Adam optimizer, and all the datasets were split for training, validation, and testing at a ratio of 6:2:2. All the models were tested over five iterations, and the average performance was recorded. Three evaluation metrics were applied to evaluate the performance of each model: mean absolute error (MAE), mean absolute percentage error (MAPE), and root mean squared error (RMSE).

### 5.5 Comparison with Baseline Methods

Table 1 compares the proposed HAGCN and benchmark models for traffic speed forecasting. TABLE I reveals that the proposed model consistently outperformed the existing state-of-the-art models on both the PeMSD4 and PeMSD8 datasets.

TABLE 1
PERFORMANCE COMPARISON OF DIFFERENT APPROACHES FOR TRAFFIC SPPED PREDICTION ON THE PEMSD4 AND PEMSD8

| Dataset | Method | Horizon 3 | | | Horizon 6 | | | Horizon 9 | | | Horizon 12 | | |
|---|---|---|---|---|---|---|---|---|---|---|---|---|---|
| | | MAE | MAPE | RMSE | MAE | MAPE | RMSE | MAE | MAPE | RMSE | MAE | MAPE | RMSE |
| PeMSD4 | DCRNN | 1.6247 | 0.0382 | 3.5105 | 1.9831 | 0.0412 | 4.0364 | 2.2974 | 0.0503 | 4.3821 | 2.4154 | 0.0529 | 4.7099 |
| | GMAN | 1.3072 | 0.0269 | 2.6123 | 1.5023 | 0.0304 | 3.3124 | 1.6842 | 0.0351 | 3.8146 | 1.8462 | 0.0405 | 4.2783 |
| | GWNET | 1.2832 | 0.0248 | 2.5261 | 1.5405 | 0.0311 | 3.2851 | 1.7442 | 0.0353 | 3.8592 | 1.9189 | 0.0399 | 4.3187 |
| | DMSTGCN | 1.2080 | 0.0234 | 2,5053 | 1.4663 | 0.0295 | 3.2636 | 1.6661 | 0.0346 | 3.8278 | 1.8356 | 0.0390 | 4.2697 |
| | ACGRN | 1.2409 | 0.0251 | 2.5817 | 1.4301 | 0.0298 | 3.1458 | 1.5627 | 0.0333 | 3.8354 | 1.6663 | 0.0360 | 3.8354 |
| | Z-GCNETs | 1.1900 | 0.0235 | 2.4342 | 1.3951 | 0.0287 | 3.0220 | 1.5364 | 0.0325 | 3.7452 | 1.6480 | 0.0355 | 3.7452 |
| | STG-NCDE | 1.1724 | 0.0221 | 2.3867 | 1.3936 | 0.0283 | 3.0545 | 1.5357 | 0.0316 | 3.4675 | 1.6756 | 0.0349 | 3.7424 |
| | **HAGCN** | **1.1291** | **0.0217** | **2.3255** | **1.3336** | **0.0270** | **2.9264** | **1.4667** | **0.0306** | **3.2949** | **1.5677** | **0.0334** | **3.5963** |
| PeMSD8 | DCRNN | 1.5743 | 0.0362 | 3.3374 | 1.6949 | 0.0392 | 3.7295 | 1.7883 | 0.0415 | 3.9987 | 1.8681 | 0.0433 | 4.2008 |
| | GMAN | 1.1098 | 0.0216 | 2.3524 | 1.2321 | 0.0246 | 2.8140 | 1.4126 | 0.0286 | 3.2865 | 1.5346 | 0.0318 | 3.6281 |
| | GWNET | 1.0659 | 0.0206 | 2.2617 | 1.2824 | 0.0259 | 2.8603 | 1.4457 | 0.0290 | 3.2938 | 1.5812 | 0.0323 | 3.6300 |
| | DMSTGCN | 1.0283 | 0.0204 | 2.2165 | 1.2401 | 0.0249 | 2.8384 | 1.3923 | 0.0283 | 3.2578 | 1.5293 | 0.0316 | 3.6118 |
| | ACGRN | 1.0464 | 0.0219 | 2.3562 | 1.2067 | 0.0265 | 2.8380 | 1.3181 | 0.0294 | 3.2166 | 1.4077 | 0.0317 | 3.4888 |
| | Z-GCNETs | 1.0246 | 0.0212 | 2.2364 | 1.2008 | 0.0256 | 2.7830 | 1.3183 | 0.0289 | 3.1636 | 1.4024 | 0.0315 | 3.4378 |
| | STG-NCDE | 0.9786 | 0.0186 | 2.0817 | 1.1812 | 0.0244 | 2.7335 | 1.3224 | 0.0288 | 3.2735 | 1.4353 | 0.0321 | 3.5042 |
| | **HAGCN** | **0.9377** | **0.0181** | **2.0494** | **1.1136** | **0.0227** | **2.6491** | **1.2281** | **0.0260** | **3.0427** | **1.3133** | **0.0284** | **3.2770** |

In this study, traffic speed prediction was used as the main task, but in studies such as Z-GCNETs and STG-NCDE, traffic volume prediction was used as the main task. Therefore, traffic volume prediction was also compared with the benchmark models and showed superior performance. This implies that our model works even on a general task. The DCRNN model is highly dependent on a predefined graph; therefore, it does not accurately capture the spatial correlation between the nodes. Therefore, its performance was slightly lower than that of the other benchmark models. In the GMAN and GraphWavenet methods, a marginally better performance was achieved because modeling using an adaptive graph was performed to precisely capture the spatial correlation between nodes that could not be captured by the DCRNN. In addition, the DMSTGCN, ACRGN, and Z-GCNET models demonstrated high performance by designing networks that effectively utilize the dynamic characteristics between nodes in different ways. STG-NCDE achieved competitive performance by applying a neural-controlled differential equation. However, the improvement gain of our model for the next most accurate method was between 4.87% and 6.35% in the MAE of horizon 12 for the PeMSD4 and PeMSD8 datasets, and our model outperformed other models over horizons 3 to 9. Our method for capturing subtle relationships by defining separate spatial dependencies for each hidden channel performed better than that achieved by providing the same graph to each channel. In addition, the method of separately calculating the importance of the adjacency matrix channel at each instance wherein time-graph convolution is performed also significantly contributed to the performance of this method. Our experiments showed that considering these factors can provide significant advantages for predicting traffic signals.

## 5.6 Evaluating Effectiveness of Key Designs

To better understand the importance of the different components of our model, we conducted ablation studies, and the corresponding results are presented in TABLE 2. The most essential part of the proposed model is the division of the spatial correlation between nodes into static and dynamic adjacency matrices and the graph channel attention module used for graph convolution. Therefore, to verify the validity of the core module, we constructed and trained three types of models and compared their results.

•without dynamic module: In this model, we trained using only the static graph module to validate the dynamic graph module.

•without static module: In this model, we trained using only the dynamic graph module to demonstrate the validity of the static graph module.

•without heterogeneous graph: This model used general graph convolution after generating a dynamic graph and static graph by changing the adjacency matrix to a homogenous graph.

•without channel attention: In this model, we used only simple graph convolution to demonstrate the importance of graph channel attention when performing graph convolution.

•without decentralization pooling layer: In this model, we do not use network decentralization pooling when using channel attention. We train with channel attention using general global average pooling.

TABLE 2
ABALATION STUDY OF THE NETWORK ARCHITECTURE

| Data set | Method | Horizon 3 | | | Horizon 6 | | | Horizon 9 | | | Horizon 12 | | |
|---|---|---|---|---|---|---|---|---|---|---|---|---|---|
| | | MAE | MAPE | RMSE | MAE | MAPE | RMSE | MAE | MAPE | RMSE | MAE | MAPE | RMSE |
| PeMSD4 | w/o dynamic | 1.2406 | 0.0251 | 2.5695 | 1.4389 | 0.0302 | 3.1200 | 1.5763 | 0.0339 | 3.4983 | 1.6814 | 0.0367 | 3.7738 |
| | w/o static | 1.2908 | 0.0272 | 2.6991 | 1.4636 | 0.0314 | 3.1733 | 1.5801 | 0.0343 | 3.4906 | 1.6701 | 0.0368 | 3.7280 |
| | w/o heterogeneous | 1.1949 | 0.0235 | 2.4694 | 1.3916 | 0.0285 | 2.9917 | 1.5387 | 0.0334 | 3.4138 | 1.6421 | 0.0356 | 3.7001 |
| | w/o channel attention | 1.2532 | 0.0258 | 2.5946 | 1.4276 | 0.0301 | 3.0904 | 1.5439 | 0.0332 | 3.4245 | 1.6328 | 0.0355 | 3.6687 |
| | w/o decentralization pooling | 1.1515 | 0.0221 | 2.3399 | 1.3479 | 0.0272 | 2.9247 | 1.4776 | 0.0308 | 3.3026 | 1.5829 | 0.0336 | 3.5812 |
| | **HAGCN** | **1.1291** | **0.0217** | **2.3255** | **1.3336** | **0.0270** | **2.9264** | **1.4667** | **0.0306** | **3.2949** | **1.5677** | **0.0334** | **3.5963** |
| PeMSD8 | w/o dynamic | 1.0267 | 0.0202 | 2.1675 | 1.2039 | 0.0247 | 2.7097 | 1.3261 | 0.0280 | 3.0806 | 1.4216 | 0.0307 | 3.3588 |
| | w/o static | 1.0541 | 0.0221 | 2.3332 | 1.1997 | 0.0287 | 2.8016 | 1.2982 | 0.0287 | 3.1212 | 1.3744 | 0.0305 | 3.3555 |
| | w/o heterogeneous | 0.9654 | 0.0183 | 2.0625 | 1.1494 | 0.0233 | 2.6636 | 1.2740 | 0.0272 | 3.0912 | 1.3677 | 0.0302 | 3.3950 |
| | w/o channel attention | 0.9932 | 0.0194 | 2.1142 | 1.1547 | 0.0242 | 2.7087 | 1.2648 | 0,0270 | 3.0813 | 1,3502 | 0.0295 | 3.3186 |
| | w/o decentralization pooling | 0.9497 | 0.0181 | 2.0460 | 1.1267 | 0.0229 | 2.6512 | 1.2461 | 0.0265 | 3.0677 | 1.3376 | 0.0291 | 3.2915 |
| | **HAGCN** | **0.9377** | **0.0181** | **2.0494** | **1.1136** | **0.0227** | **2.6491** | **1.2281** | **0.0260** | **3.0427** | **1.3133** | **0.0284** | **3.2770** |

The results show that all the major modules contribute to the learning outcomes of the model. Learning with the dynamic module alone achieved a better performance than learning with only the static module. However, we observed that the performance of the model improved significantly only when two modules were used simultaneously. In addition, using the channel attention module for graph convolution outperformed the model without attention, with a relative gain of 0.6% to 0.9% in MAE for each dataset. This indicates the importance of modeling the attention of adjacency matrix channels when using heterogeneity-aware graph neural networks. Moreover, using the network decentralization-based attention module, MAE has a relative gain between 1.0% and 1.8% for each data set compared to the global average pooling module. These results mean that giving attention to the proposed decentralization module significantly affects the results of the prediction model.

### 5.7 Static/Dynamic Module Methodology and Traffic Speed Forecasting Visualization

Fig. 4 shows a part of an adjacency matrix trained with a static/dynamic generator trained on the PeMSD4 dataset. The heatmaps in Fig. 4 were trained with the same initial weight, but compared to the completed heatmap, they were trained with completely different weights. It can be seen that the spatial correlation between each sensor was well trained by dividing it into a static relationship and dynamically changing relationship.

We visualized and compared the results for some nodes on the PeMSD4 test dataset. In Fig. 5, it can be observed that the prediction results of our proposed model estimate the true value better. This is the result of HAGCN organically using information from other nodes when performing traffic forecasting.

### 5.8 How did heterogeneity channel attention work?

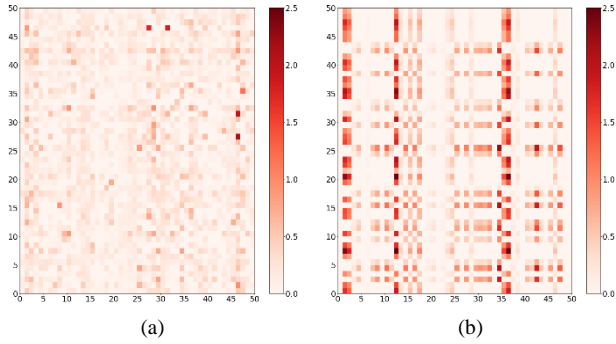

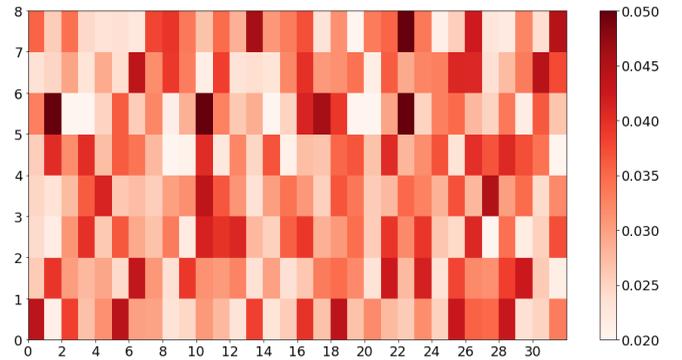

Fig 4. (a) Heatmap of the learned static module's adjacency matrix for the first 50 nodes in the first channel (b) Time-averaged heatmap of the adjacency matrix of dynamic module for the first 50 nodes

Fig 6. Channel attention values in 8 HAGCN blocks of the static module's adjacency matrix

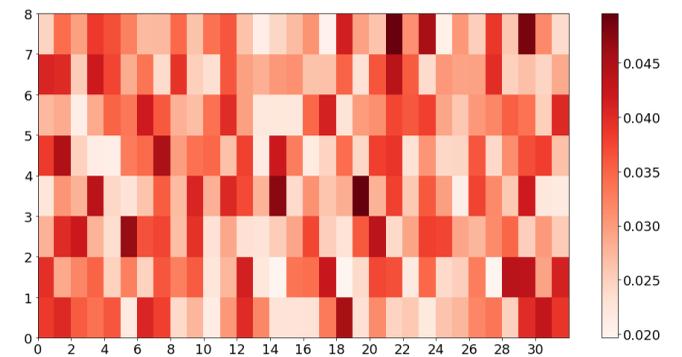

Fig 7. Channel attention values in 8 HAGCN blocks of the dynamic module's adjacency matrix

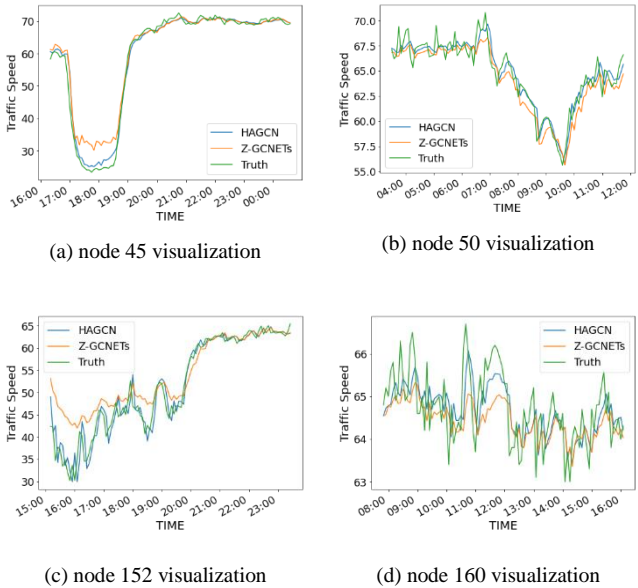

Fig 5. Traffic speed forecasting visualization example

Fig. 6 and Fig. 7 are heatmaps of channel attention values of all HAGCN blocks in the static/dynamic module. The x-axis means each channel and the y-axis means each HAGCN block. Each time graph convolution occurs, the importance of each hidden channel changes. This is because the characteristic of the hidden feature that is calculated every time a temporal relationship is learned from the HAGCN block to the TCN changes. Since we set the number of hidden channels to 32 when learning the PeMSD4 dataset, if the attention values of all channels are the same or similar, it should have a value close to 1/32. However, in Fig. 6 and Fig. 7, whenever our model performs graph convolution, it captures changes in channel attention well and has various attention values.

### 5.9 Traffic Volume Task Comparision

To examine whether the HAGCN model can predict other traffic signals in addition to the traffic speed prediction task, we have also experimented with the traffic volume prediction in the PeMSD dataset. The experimental settings for the traffic volume prediction task were the same, except that the initial learning rate was 0.003 and the size of the output channels of the dilated and graph convolutions was set to 64. All the models were tested for five iterations, and the average performance was recorded in TABLE 3. The improvement gain of our model compared to the second most accurate method, the STG-NCDE model, was between 2.65% and 3.88% in the MAE of horizon 12 for the PeMSD4 and PeMSD8 datasets, respectively.

TABLE 3
PERFORMANCE COMPARISON FOR TRAFFIC VOLUME PREDICTION ON THE PEMSD4 AND PEMSD8 DATASETS

| Model | PeMSD4 | | | PeMSD8 | | |
|---|---|---|---|---|---|---|
| Horizon 12 | MAE | RMSE | MAPE | MAE | RMSE | MAPE |
| Graph WaveNet | 24.89 | 39.66 | 0.1729 | 18.28 | 30.08 | 0.1215 |
| AGCRN | 19.83 | 32.26 | 0.1297 | 15.95 | 25.22 | 0.1009 |
| Z-GC NETs | 19.50 | 31.61 | 0.1278 | 15.76 | 25.11 | 0.1001 |
| STG-NCDE | 19.21 | 31.09 | 0.1276 | 15.45 | 24.81 | 0.0990 |
| **HAGCN** | **18.70** | **30.61** | **0.1260** | **14.85** | **23.87** | **0.0957** |

### 5.10 Robustness Check

To test the robustness of our model under noisy conditions, it was trained with Gaussian noise. The added noise followed a Gaussian density with zero mean and a variance of 2,4. TABLE 4 presents an MAE comparison with the benchmark model with two different levels of noise training. It can be seen that our model consistently outperformed the other models on both the PeMSD4 and PeMSD8 datasets when noise was added. Moreover, compared to other methods, it can be seen that our model's performance deteriorates slightly when Gaussian noise is added gradually.

TABLE 4
ROBUSTNESS CHECK

| Dataset | Noise | AGCRN | Z_GCNETs | HAGCN |
|---|---|---|---|---|
| PeMSD4 | X | 1.6663 | 1.6480 | **1.5677** |
| | N(0,2) | 1.7140 | 1.7118 | **1.6497** |
| | N(0,4) | 1.7984 | 1.7957 | **1.7204** |
| PeMSD8 | X | 1.4077 | 1.4024 | **1.3133** |
| | N(0,2) | 1.5129 | 1.4856 | **1.4010** |
| | N(0,4) | 1.5769 | 1.5709 | **1.4607** |

## 6 CONCLUSION

In this study, we presented a novel framework for modeling spatiotemporal graphs. By dividing the spatial relationship between the sensors into two static/dynamic relationships, we proposed a generator that creates a heterogeneous graph because the spatial dependency of each hidden channel is different. In addition, using the advantage of a heterogeneous graph, we proposed an effective framework for learning spatial dependency by automatically learning the importance of each hidden channel during graph convolution. Our model achieves state-of-the-art results for two sets of traffic data. We consider that the use of a GCN by appropriately generating heterogeneity-aware graphs utilizing hidden channels is a promising method for spatiotemporal processing and improves prediction performance.

## REFERENCES

[1] Z. Lv, J. Xu, K. Zheng, H. Yin, P. Zhao, and X. Zhou, "Lc-rnn: A deep learning model for traffic speed prediction," in *IJCAI*, 2018, pp. 3470-3476.

[2] C. Zheng, X. Fan, C. Wen, L. Chen, C. Wang, and J. Li, "Deepstd: Mining spatio-temporal disturbances of multiple context factors for citywide traffic flow prediction," *IEEE Transactions on Intelligent Transportation Systems*, vol. 21, no. 9, pp. 3744-3755, 2019.

[3] B. Yu, H. Yin, and Z. Zhu, "Spatio-Temporal Graph Convolutional Networks: A Deep Learning Framework for Traffic Forecasting," in *IJCAI*, 2018.

[4] Y. Li, R. Yu, C. Shahabi, and Y. Liu, "Diffusion Convolutional Recurrent Neural Network: Data-Driven Traffic Forecasting," in *International Conference on Learning Representations*, 2018.

[5] Z. Wu, S. Pan, G. Long, J. Jiang, and C. Zhang, "Graph WaveNet for Deep Spatial-Temporal Graph Modeling," in *The 28th International Joint Conference on Artificial Intelligence (IJCAI)*, 2019: International Joint Conferences on Artificial Intelligence Organization.

[6] C. Zheng, X. Fan, C. Wang, and J. Qi, "Gman: A graph multi-attention network for traffic prediction," in *Proceedings of the AAAI Conference on Artificial Intelligence*, 2020, vol. 34, no. 01, pp. 1234-1241.

[7] L. Han, B. Du, L. Sun, Y. Fu, Y. Lv, and H. Xiong, "Dynamic and Multi-faceted Spatio-temporal Deep Learning for Traffic Speed Forecasting," in *Proceedings of the 27th ACM SIGKDD Conference on Knowledge Discovery & Data Mining*, 2021, pp. 547-555.

[8] L. BAI, L. Yao, C. Li, X. Wang, and C. Wang, "Adaptive Graph Convolutional Recurrent Network for Traffic Forecasting," *Advances in Neural Information Processing Systems,* vol. 33, 2020.

[9] Y. Chen, I. Segovia-Dominguez, and Y. R. Gel, "Z-GCNETs: Time Zigzags at Graph Convolutional Networks for Time Series Forecasting," *arXiv preprint arXiv:2105.04100*, 2021.

[10] B. M. Williams and L. A. Hoel, "Modeling and forecasting vehicular traffic flow as a seasonal ARIMA process: Theoretical basis and empirical results," *Journal of transportation engineering,* vol. 129, no. 6, pp. 664-672, 2003.

[11] E. Zivot and J. Wang, "Vector autoregressive models for multivariate time series," *Modeling Financial Time Series with S-Plus®,* pp. 385-429, 2006.

[12] S. Hochreiter and J. Schmidhuber, "Long short-term memory," *Neural computation,* vol. 9, no. 8, pp. 1735-1780, 1997.

[13] J. Chung, C. Gulcehre, K. Cho, and Y. Bengio, "Empirical evaluation of gated recurrent neural networks on sequence modeling," in *NIPS 2014 Workshop on Deep Learning, December 2014*, 2014.

[14] R. Sen, H.-F. Yu, and I. S. Dhillon, "Think globally, act locally: A deep neural network approach to high-dimensional time series

forecasting," *Advances in neural information processing systems,* vol. 32, 2019.

[15] S. Bai, J. Z. Kolter, and V. Koltun, "Trellis Networks for Sequence Modeling," in *International Conference on Learning Representations*, 2018.

[16] Z. Lv, J. Xu, K. Zheng, H. Yin, P. Zhao, and X. Zhou, "Lc-rnn: A deep learning model for traffic speed prediction," in *IJCAI*, 2018, vol. 2018, p. 27th.

[17] Q. Zhang, J. Chang, G. Meng, S. Xiang, and C. Pan, "Spatio-temporal graph structure learning for traffic forecasting," in *Proceedings of the AAAI Conference on Artificial Intelligence*, 2020, vol. 34, no. 01, pp. 1177-1185.

[18] C. Song, Y. Lin, S. Guo, and H. Wan, "Spatial-temporal synchronous graph convolutional networks: A new framework for spatial-temporal network data forecasting," in *Proceedings of the AAAI Conference on Artificial Intelligence*, 2020, vol. 34, no. 01, pp. 914-921.

[19] R. Huang, C. Huang, Y. Liu, G. Dai, and W. Kong, "LSGCN: Long Short-Term Traffic Prediction with Graph Convolutional Networks," in *IJCAI*, 2020, pp. 2355-2361.

[20] M. Li and Z. Zhu, "Spatial-Temporal Fusion Graph Neural Networks for Traffic Flow Forecasting," in *Proceedings of the AAAI Conference on Artificial Intelligence*, 2021, vol. 35, no. 5, pp. 4189-4196.

[21] S. Yang, J. Liu, and K. Zhao, "Space Meets Time: Local Spacetime Neural Network For Traffic Flow Forecasting," in *2021 IEEE International Conference on Data Mining (ICDM)*, 2021: IEEE, pp. 817-826.

[22] J. Choi, H. Choi, J. Hwang, and N. Park, "Graph Neural Controlled Differential Equations for Traffic Forecasting," *arXiv preprint arXiv:2112.03558,* 2021.

[23] T. G. Kolda and B. W. Bader, "Tensor decompositions and applications," *SIAM review,* vol. 51, no. 3, pp. 455-500, 2009.

[24] J. Hu, L. Shen, and G. Sun, "Squeeze-and-excitation networks," in *Proceedings of the IEEE conference on computer vision and pattern recognition*, 2018, pp. 7132-7141.

[25] J. Wang, Y. Chen, R. Chakraborty, and S. X. Yu, "Orthogonal convolutional neural networks," in *Proceedings of the IEEE/CVF conference on computer vision and pattern recognition*, 2020, pp. 11505-11515.

[26] L. C. Freeman, "Centrality in social networks conceptual clarification," *Social networks,* vol. 1, no. 3, pp. 215-239, 1978.

[27] A. Van Den Oord *et al.*, "WaveNet: A generative model for raw audio," *SSW*, vol. 125, p. 2, 2016.

[28] C. Chen, K. Petty, A. Skabardonis, P. Varaiya, and Z. Jia, "Freeway performance measurement system: mining loop detector data," *Transportation Research Record,* vol. 1748, no. 1, pp. 96-102, 2001.